\documentclass[a4paper,12pt]{article}
\usepackage[utf8]{inputenc}
\usepackage{threeparttable}
\pdfoutput=1
\usepackage{verbatim}
\usepackage{epsfig}
\usepackage{epsf}
\usepackage{epstopdf}
\usepackage{graphicx}
\usepackage{textcomp}
 \usepackage{amsmath}
\usepackage{amssymb}
\usepackage{color}
\definecolor{Blue}{rgb}{0.3,0.3,0.9}
\definecolor{red}{rgb}{1,0,0}
\usepackage{rotating}
\usepackage{pgfplots}
\usepackage{tikz}
\usepackage{url}
\usetikzlibrary{patterns}
\usepackage{amsfonts,amsthm,titling,url,array,mathtools}
\usepackage[superscript,biblabel]{cite}
\usepackage[toc,acronym]{glossaries}

\title{Generative Adversarial Networks Synthesize Realistic OCT Images of the Retina}
\author{Stephen G. Odaibo,\\ M.D.,M.S.(Math),M.S.(Comp. Sci.)\\{\tiny{.}}\\\vspace{10pt}RETINA-AI Health, Inc.}

\begin{document}

\maketitle
\thispagestyle{empty}
\begin{abstract}
We report, to our knowledge, the first end-to-end application of Generative Adversarial Networks (GANs) towards the synthesis of Optical Coherence Tomography (OCT) images of the retina. Generative models have gained recent attention for the increasingly realistic images they can synthesize, given a sampling of a data type. In this paper, we apply GANs to a sampling distribution of OCTs of the retina. We observe the synthesis of realistic OCT images depicting recognizable pathology such as macular holes, choroidal neovascular membranes, myopic degeneration, cystoid macular edema, and central serous retinopathy amongst others. This represents the first such report of its kind. Potential applications of this new technology include for surgical simulation, for treatment planning, for disease prognostication, and for accelerating the development of new drugs and surgical procedures to treat retinal disease.

\vspace{135pt}{\hspace{-17pt}\small{\textbf{Correspondence Email:}\\\hspace{10pt} stephen.odaibo@retina-ai.com}}
\end{abstract}

\newpage
\section{Introduction}

Optical Coherence Tomography (OCT)\cite{husw1991} has revolutionized the field of retina. And usage of this technology has become standard in the management of the vast majority of retinal patient encounters. An understanding of the native probability distribution of OCT representation of retinal disease is therefore essential. This is especially so as we seek to acquire more personalized understanding of disease, so as to develop treatments that are most fitting to the pathology of interest.

Following the work of Goodfellow et al\cite{gopo2014}, Generative Adversarial Networks (GANs), an adversarial type of generative model, has gained popularity based on the increasingly realistic datasets it can generate. Some theory of adversarial training algorithms has been in existence for at least 3 decades (Schmidhuber, 1992)\cite{sc1992}. However, practical feasibility, awareness, traction, and performance are only now notably rising. Interesting applications have included "deep fakes" of human faces, and more recently, of medical data. For instance, Burlina et al \cite{bujo2019} recently used GAN architecture to synthesize fundus images of age-related macular degeneration.

To our knowledge, this is the first use of generative adversarial networks to perform end-to-end synthesis OCT images of the retina. GANs have recently been used for the segmentation of various anatomical parts of the eye within ophthalmic images~\cite{tego2018,lifu2018}, as well as conditional adversarial generation from vessel trees~\cite{coga2018}. They have also been used to generate fundus photographs of the retina~\cite{dico2018}.

The capability to sample never before seen or even not yet existing retinal pathology will increase both our understanding of disease and facilitate development and assessment of new therapies. One such pathological feature which our GAN architecture detected is macular edema as well as subretinal fluid. Firstly, the presence of macular edema or subretinal fluid in the macula is essentially always indicative
of pathology. It is a canonical hallmark that underlies several common retinal conditions such as diabetic macular edema\cite{hepu1998,madu2003,hepu1995,otki1999,puhe1995}, exudative macular degeneration
\cite{heba1996,husw1991,swiz1993,coco2007,kepa2012}, retinal vein occlusions \cite{rofu2005}, pseudophakic macular edema \cite{vooh2007,sosa1999,kibe2008,kieq2007,peut2007}
central serous chorioretinopathy \cite{hepu1995b,iiha2000,beka2000,moru2005,fugo2008}, and macula-off retinal detachments \cite{wogo2002,bahi2004,besc2007,wo2004,naha2009}. And therefore having a fuller understanding of all the ways and forms of macular edema will be essential to advancing our ability to develop effective therapies for a great number of retinal diseases.

In the remainder of this paper, we present our methods, results, and a discussion of the implications of this development for retinal research.

\section{Methods}

\begin{figure}[h]
\begin{center}
\scalebox{.40}
{\includegraphics{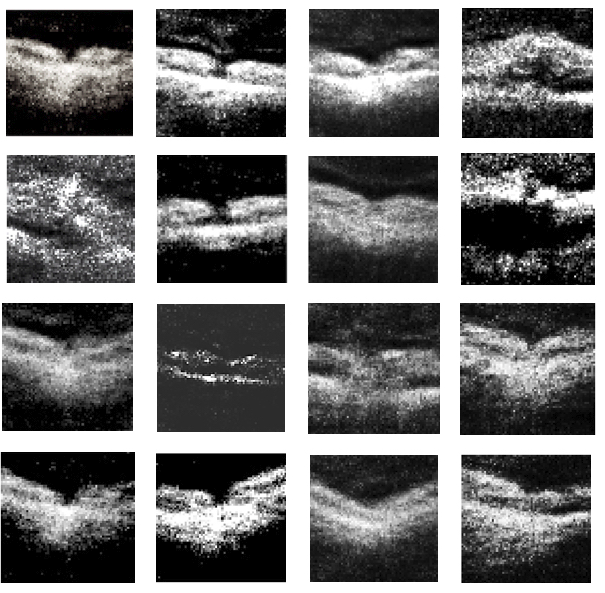}}
\end{center}
\caption[AI99800]{AI Generated images. Sampled from Epoch 96000 to 99800}
\label{fig:AI-images}
\end{figure}

\begin{figure}[h]
\begin{center}
\scalebox{.40}
{\includegraphics{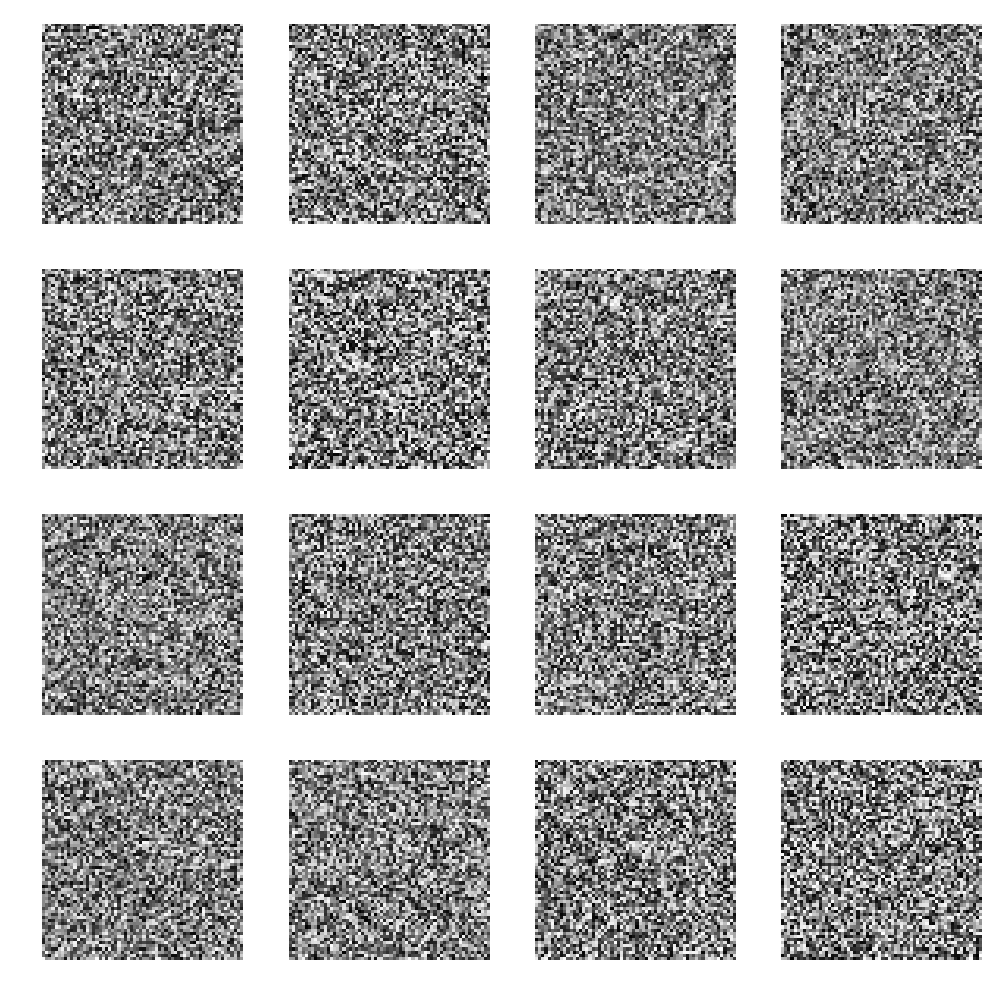}}
\end{center}
\caption[Deconv]{Random Deconvolution output. Epoch 0}
\label{fig:Generator initial output}
\end{figure}

\begin{figure}[h]
\begin{center}
\scalebox{.40}
{\includegraphics{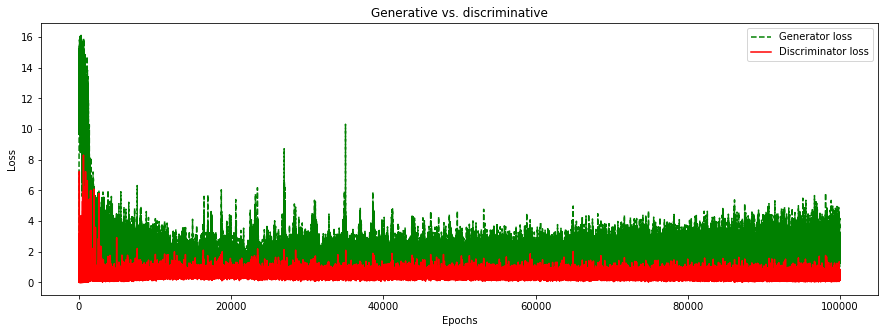}}
\end{center}
\caption[all]{Generator and Discriminator losses: All epochs}
\label{fig:GvsD all}
\end{figure}

\begin{figure}[h]
\begin{center}
\scalebox{.40}
{\includegraphics{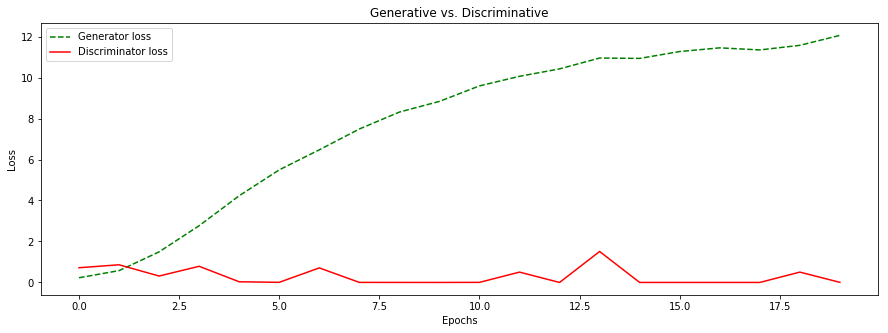}}
\end{center}
\caption[mid]{Generator and Discriminator losses between Epoch 0 and 20}
\label{fig:GvsD onset}
\end{figure}

\begin{figure}[h]
\begin{center}
\scalebox{.40}
{\includegraphics{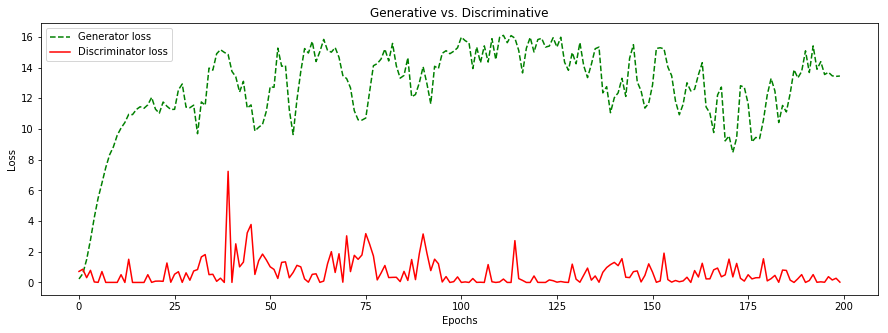}}
\end{center}
\caption[Earlymid]{Generator and Discriminator losses between Epoch 0 and 200}
\label{fig:GvsD early to mid}
\end{figure}

\begin{figure}[h]
\begin{center}
\scalebox{.40}
{\includegraphics{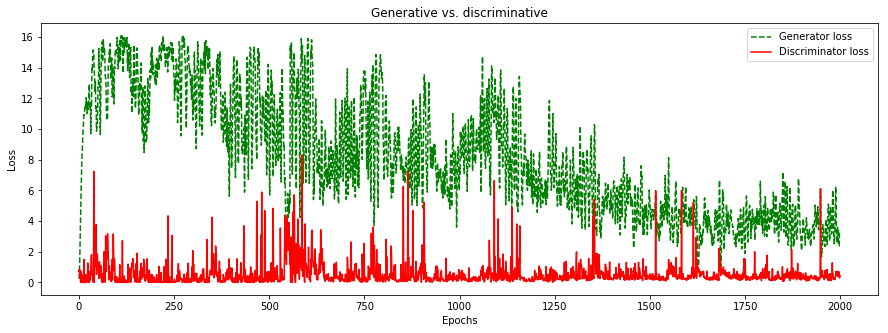}}
\end{center}
\caption[mid]{Generator and Discriminator losses between Epoch 0 and 2000}
\label{fig:GvsD early}
\end{figure}

\begin{figure}[h]
\begin{center}
\scalebox{.40}
{\includegraphics{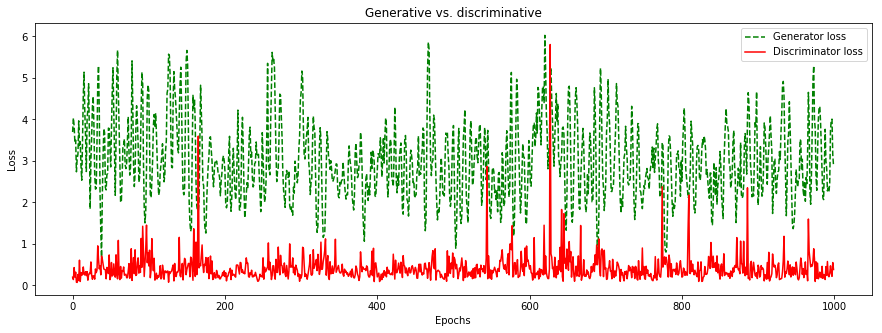}}
\end{center}
\caption[mid]{Generator and Discriminator losses between Epoch 2000 and 3000}
\label{fig:GvsD mid}
\end{figure}

\begin{figure}[h]
\begin{center}
\scalebox{.40}
{\includegraphics{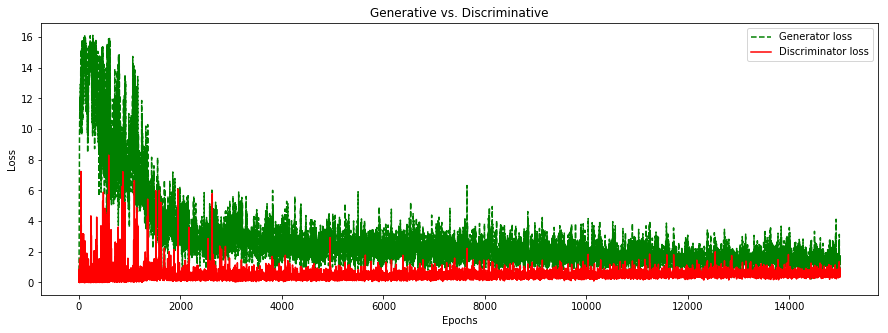}}
\end{center}
\caption[Earlymid2]{Generator and Discriminator losses between Epoch 0 and 15000}
\label{fig:GvsD early mid}
\end{figure}

A Generative Adversarial Network was implemented in Keras. The algorithm was run on an 8 core NVIDIA V100 machine equivalent with 600 GB of RAM. A database consisting of 500,000 images after augmentation was utilized for the training. The database had a diverse representation of pathology including macular holes, cystoid macular edema, exudative and non-exudative macular degeneration, central serous retinopathy, macula-off retinal detachments, and normal retina. Training was done for 100,000 epochs. A randomized vector was deconvolved according to the architecture depicted in Figure 1, yielding the generator output. The deconvolutional neural network weights were iteratively trained via backpropagation as directed by its loss function, a binary cross entropy. Our GAN architecture had some similarity as well as some notable differences compared to that recommended heuristically by Radford et al~\cite{rame2015}. LeakyReLu was used for the activation function of all layers of both the generator and discrimator, except for their final layer where a tanh and a sigmoid was used respectively. Batchnormalization was done on every layer of the generator and on no layer of the discriminator. 

\section{Results}

Figure~\ref{fig:AI-images} shows realistic OCT images of the retina which were synthesized after 99800 training epochs of the algorithm. At the onset, prior to training, the initial random vector of size 100 yielded Figure~\ref{fig:Generator initial output} from the generator. Figure~\ref{fig:GvsD mid} to Figure~\ref{fig:GvsD early mid} depict the loss landscape of the discriminator and generator during training. 

At the onset of training, where the discriminator is first getting trained, we see a decrease in the discriminator's loss and a concurrent rise in the generator's loss. This early behavior is even marked by an early crossing in the loss graphs, as depicted in Figure~\ref{fig:GvsD onset} and Figure~\ref{fig:GvsD early}. Of note, in the loss graphs depicted in Figures~\ref{fig:GvsD all}, \ref{fig:GvsD early}, and \ref{fig:GvsD early to mid}, we see an asymptotic convergence (within some neighborhood) of the losses of both the generator and discriminator towards zero. This however does not by itself constitute completion of the training process. The persistent fluctuations in both discriminator and generator loss graphs are indicators of the continuing adversarial contest that continues well past the point at which the graphs appear to have settled close to each other. Figure~\ref{fig:GvsD mid} clearly shows the adversariality between the discriminator and the generator. As the discriminator loss rises, the generator loss falls, and vice versa.

\section{Discussion}

Our results demonstrate that realistic OCT images of the retina can be generated using Generative Adversarial Networks. This has many potential applications in the field of ophthalmology and in medicine and healthcare in general. In the current paradigm, physicians are trained by seeing cases of patients with certain diseases. The principles governing the diagnosis and management of such patients are learned during care. However, for less common or rare diseases, physicians in training get very limited exposure, and as such limited training. Our GAN approach points to a means of augmenting case loads and experience where needed. Another potential application for this is in the simulation of patient responses to certain drugs or surgical procedures in development. With a realistic model of the human retina, one is able to perform increasingly realistic simulations. This holds promise to significantly accelerate the pace of research, discovery, and development of therapies.

Unlike with traditional supervised learning problems such as image classification, the endpoint at which training can be halted is less clear. It is a subjective decision based on similarity of the generated images to the training sample distribution which it seeks to sample. This implies a need for the programmer to have an understanding of the data and the native probability distribution of that data type. Hence domain specific knowledge and interdisciplinarity will be necessary requisites for moving this field of research forward.

\section{Conclusion}
Here, we presented the first reported application of generative adversarial networks for the synthesis of OCT images of the retina. Our approach yielded realistic images depicting various retinal diseases such as macular holes, exudative macular degeneration, pathological myopic myopic, central serous retinopathy, and macula-off retinal detachments amongst others. This result opens up an avenue to several applications including for the acceleration of pharmacological and surgical therapies. 

\section*{Acknowledgement}
The author thanks Daniel T. Chang of IBM (retired) for endorsement to the arXiv's Computer Science Artificial Intelligence Section. And he thanks Google Cloud Program for Startups for providing the computational resources to RETINA-AI Health, Inc for this study.

\bibliographystyle{plain}
\bibliography{/home/sodaibo/Documents/BOOKS/mybibliography_2015_nov}

\begin{figure}[h]
\begin{center}
\scalebox{.40}
{\includegraphics*{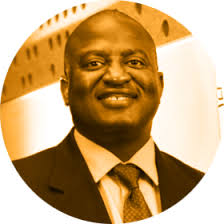}}
\end{center}
\caption{\textbf{About the Author:}  Dr. Stephen G. Odaibo is Founder, CEO, and Chief Software Architect of RETINA-AI Health Inc, a company using Artificial Intelligence to improve Healthcare. He is a Retina specialist, Mathematician, Computer Scientist, and Full-Stack AI Engineer. Dr. Odaibo is the only Ophthalmologist in the world with advanced degrees in both Mathematics and Computer Science. In 2017 UAB College of Arts and Sciences awarded Dr. Odaibo its highest honor, the Distinguished Alumni Achievement Award. In 2005 he won the Barrie Hurwitz Award for Excellence in Clinical Neurology at Duke Univ. School of Medicine where he topped the class in Neurology and in Pediatrics. In 2016 Dr. Odaibo delivered the Opening Keynote address at the Global Ophthalmologists Meeting in Osaka Japan. And he delivered the inaugural Special Guest Lecture in Ophthalmology at the University of Ilorin, Nigeria. In 2018, Dr. Odaibo delivered the keynote address at the National Medical Association's New Innovations in Ophthalmology Session. And he delivered a Plenary Keynote address on AI in Healthcare at AI Expo Africa in Cape town, South Africa. He is author of the book "Quantum Mechanics and the MRI Machine'' (2012), and of the book "The Form of Finite Groups: A Course on Finite Group Theory" (2016).}
\label{fig:auth}
\end{figure}

\end{document}